\documentclass[10pt,twocolumn,letterpaper]{article}

\usepackage{iccv}
\usepackage{times}
\usepackage{epsfig}
\usepackage{graphicx}
\usepackage{amsmath}
\usepackage{amssymb}

\usepackage{multirow}
\usepackage{amsthm}
\usepackage{booktabs}

\usepackage{makecell}  
\usepackage{scalerel} 
\usepackage{diagbox} 
\usepackage[ruled,vlined]{algorithm2e}
\usepackage{subfig} 
\usepackage[export]{adjustbox} 
\usepackage{wrapfig} 
\usepackage{placeins}  
\usepackage{MnSymbol}

\usepackage[utf8]{inputenc}

\usepackage[pagebackref=true,breaklinks=true,colorlinks,bookmarks=false]{hyperref}

\iccvfinalcopy 

\def\iccvPaperID{13} 

\ificcvfinal\fi
\DeclareUnicodeCharacter{2212}{-}
\begin{document}

\title{Maximally Compact and Separated Features with Regular Polytope Networks}

\author{Federico Pernici, Matteo Bruni, Claudio Baecchi and Alberto Del Bimbo \\
MICC -- Media Integration and Communication Center\\
University of Florence -- Italy\\
{\tt\small \{federico.pernici, matteo.bruni, claudio.baecchi, alberto.delbimbo\}@unifi.it}
}

\maketitle

\begin{abstract}
Convolutional Neural Networks (CNNs) trained with the Softmax loss are widely used classification models for several vision tasks. Typically, a learnable transformation (i.e. the classifier) is placed at the end of such models returning class scores that are further normalized into probabilities by Softmax. This learnable transformation has a fundamental role in determining the network internal feature representation.

In this work we show how to extract from CNNs features with the properties of \emph{maximum} inter-class separability and \emph{maximum} intra-class compactness by setting the parameters of the classifier transformation as not trainable (i.e. fixed). We obtain features similar to what can be obtained with the well-known ``Center Loss'' \cite{wen2016discriminative} and other similar approaches but with several practical advantages including maximal exploitation of the available feature space representation, reduction in the number of network parameters, no need to use other auxiliary losses besides the Softmax.    

Our approach unifies and generalizes into a common approach two apparently different classes of methods regarding: discriminative features, pioneered by the Center Loss \cite{wen2016discriminative} and fixed classifiers, firstly evaluated in \cite{hoffer2018fix}.

Preliminary qualitative experimental results provide some insight on the potentialities of our combined strategy.
\end{abstract}

\section{Introduction}
Convolutional Neural Networks (CNNs) together with the Softmax loss have achieved remarkable successes in computer vision, improving the state of the art in image classification tasks  \cite{krizhevsky2012imagenet,he2016deep,Densenet2017,Zoph_2018_CVPR}. In classification all the possible categories of the test samples are also present in the training set and the predicted labels determine the performance. As a result, the Softmax with Cross Entropy loss is widely adopted by many classification approaches due to its simplicity, good performance and probabilistic interpretation.
In other applications like face recognition \cite{Cao18} or human body reidentification \cite{kalayeh2018human} test samples are not known in advance and recognition at test time is performed according to learned features based on their distance.

\begin{figure}[t]
\includegraphics[width=0.99\columnwidth]{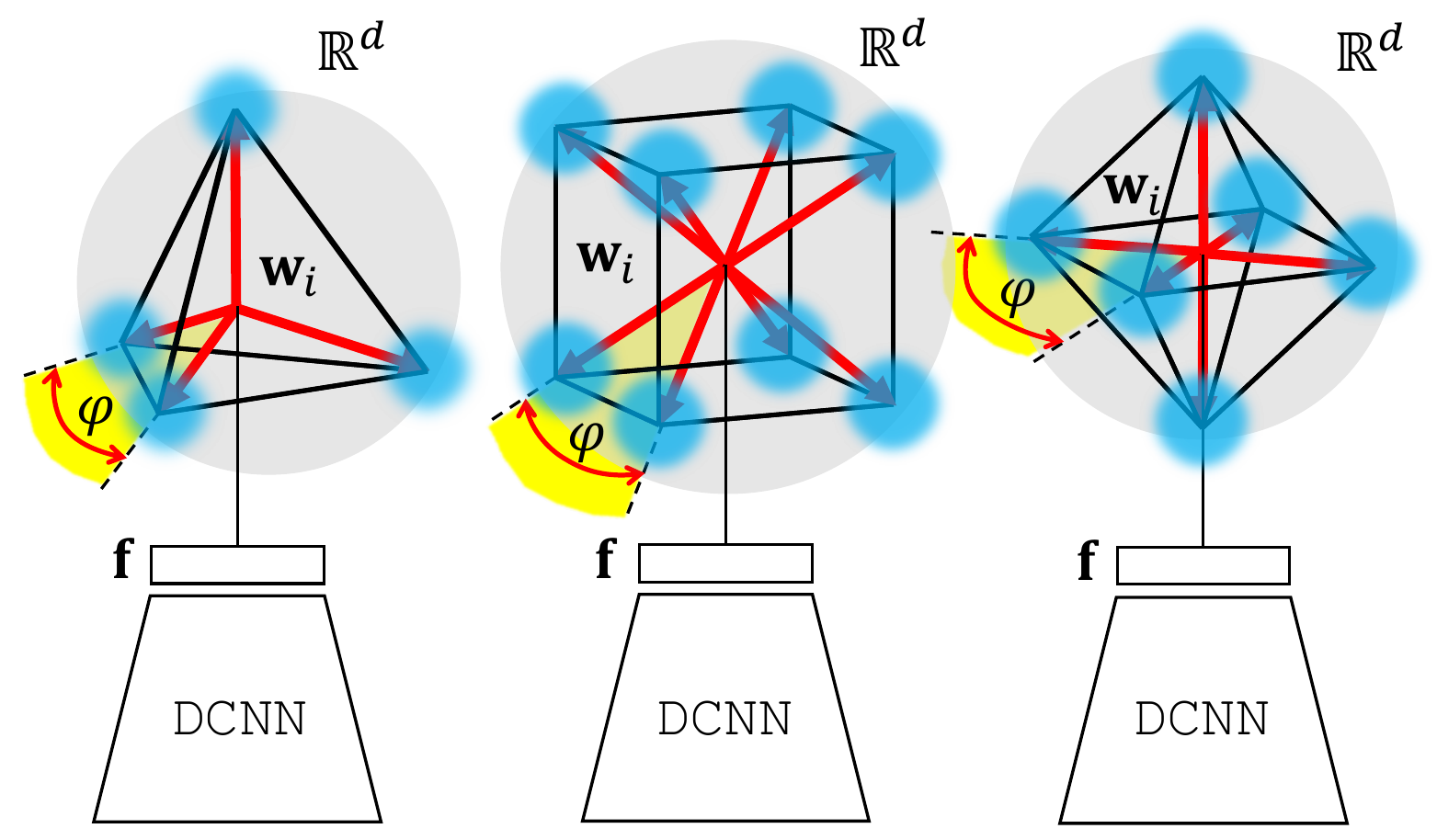}
\caption{ Margin Regular Polytope Networks (Margin-RePoNets). Features with \emph{maximal} inter-class separability and intra-class compactness are shown (light blue). These are determined combining fixed classifiers derived from regular polytopes  \cite{pernici2019fix} with a recently developed margin loss \cite{deng2019arcface}.
Maximal features separation is obtained by setting
the classifier weights $\mathbf{w}_i$ according to values following the 
symmetrical of configuration regular polytopes (red). Maximal compactness is obtained by setting the margin between the features at the maximum allowed (i.e. $\varphi$). }
\label{fig_IntroFig}
\end{figure}

The underlying assumption in this learning scenario is that images of the same identity (person) are expected to be closer in the representation space, while different identities are expected to be far apart. Or equivalently, the learned features having low intra-class distance and large inter-class distance are successful at modeling novel unseen identities and for this reason such features are typically defined ``discriminative''.
Specifically, the Center Loss, firstly proposed in \cite{wen2016discriminative}, has been proved to be an effective method to compute discriminative features. The method learns a center determined as the average of features belonging to the same class. During training, the centers are updated by minimizing the distances between the deep features and their corresponding class centers. The CNN is trained under the joint supervision of the Softmax loss and the Center Loss by balancing the two supervision signals. Intuitively, the Softmax loss forces the deep features of different classes to be separable while the Center Loss attracts the features of the same class to their centers achieving compactness.

Despite its usefulness, the Center Loss has some limitations: the feature centers are extra parameters stored outside the network that are not jointly optimized with the network parameters. Indeed, they are updated with an autoregressive mean estimator that tracks the underlying representation changes at each step. Moreover, when a large number of classes must be learned, mini-batches do not provide enough samples for a correct estimation of the mean. 
Center Loss also requires a balancing between the two supervision losses which typically requires a search over the balancing hyper-parameter. 

Some works have successfully addressed all the issues described above importing intra-class feature compactness directly into the Softmax loss. This class of methods, including \cite{ranjan2017l2,wang2017normface,deng2019arcface,wang2018cosface,Liu2017CVPR}, avoids the need of an auxiliary loss (as in the Center Loss) with the possibility of including a margin between the class decision boundaries, all in a single Softmax loss.

Other successful works follow a nearly opposite strategy by removing the final classification layer and learn directly a distance evaluated on image pairs or image triplets as shown in \cite{chopra2005learning} and in  \cite{schroff2015facenet} respectively. Despite the performance results, carefully designed pair and triplet selection is required to avoid slow convergence and instability. 

Except for few recent cases \cite{kaizhao2019regular,HypersphericalEnergy2018,pernici2019fix} inter-class separability and compactness are always enforced in a local manner without considering global inter-class separability and intra-class compactness. For this purpose, the work \cite{HypersphericalEnergy2018} uses an auxiliary loss for enforcing global separability. The work 
\cite{kaizhao2019regular} use an auxiliary loss similar to \cite{HypersphericalEnergy2018} for enforcing global separability and a further margin loss to enforce compactness. 
The work \cite{pernici2019fix} uses a fixed classifier in which the parameters of the final transformation implementing the classifier are \emph{not} subjected to learning and are set with values taken from coordinate vertices of a regular polytope. This avoids optimizing for maximal separation as in 
\cite{kaizhao2019regular} and \cite{HypersphericalEnergy2018} 
since regular polytopes naturally provide 
distributed vertices (i.e. the classifier weights) at equal angles maximizing the available space.

In this paper we address \emph{all} those limitations including global inter-class separability and compactness in a maximal sense without the need of any auxiliary loss. This is achieved by exploiting the Regular Polytope fixed classifiers (RePoNets) proposed in \cite{pernici2019fix} and improving their feature compactness according to the additive angular margin described in \cite{deng2019arcface}.   
As illustrated in Fig.~\ref{fig_IntroFig}, the advantage of the proposed combination is the capability of generating global maximally separated and compact features (shown in light blue) angularly centered around the vertices  of polytopes (i.e. the classifier fixed weights shown in red).    
The same figure further illustrates the three basic types of features that can be learned. 
Although, there are infinite regular polygons in $\mathbb{R}^2$ and 5 regular polyedra in $\mathbb{R}^3$, there are only three regular polytopes in $\mathbb{R}^d$ with $d \geq 5$, namely the $d$-Simplex, the $d$-Cube and the $d$-Orthoplex.

In particular, the angle $\varphi$ subtended between a class weight and its connected class weights is constant and maximizes inter-class separability in the available space. 
The angle $\varphi$ is further exploited 
to obtain the maximal compactness by setting the angular margin between the features to $\varphi$ (i.e. the maximum allowed margin). The advantage of our formulation is that the margin is no longer an hyperparameter that have to be searched since it is obtained from a closed form solution. 

\section{Related Work}

\textbf{Fixed Classifiers}. 
Empirical evidence, reported in \cite{hardt2016identity}, firstly shows that a CNN with a fixed classification layer does not worsen the performance on the CIFAR10 dataset. A recent paper \cite{hoffer2018fix} explores in more detail the idea of excluding the classification parameters from learning. The work shows that a fixed classifier causes little or no reduction in classification performance for common datasets (including ImageNet) while allowing a noticeable reduction in trainable parameters, especially when the number of classes is large. Setting the last layer as not trainable also reduces the computational complexity for training as well as the communication cost in distributed learning. The described approach sets the classifier with the coordinate vertices of orthogonal vectors taken from the columns of the Hadamard\footnote{The Hadamard matrix is a square matrix whose entries are either +1 or −1 and whose rows are mutually orthogonal.} matrix. Although the work uses a fixed classifier, the properties of the generated features are not explored. A major limitation of this method is that, when the number of classes is greater than the dimension of the feature space, it is not possible to have mutually orthogonal columns and therefore some of the classes are constrained to lie in a common subspace causing a reduction in classification performance. 

Recently \cite{pernici2019fix} improves in this regard showing that a novel set of unique directions taken from regular polytopes overcomes the limitations of the Hadamard matrix. The work further shows that the generated features are stationary at training time and coincide with the equiangular spaced vertices of the polytope. 
Being evaluated for classification the method does not enforce feature compactness. We extend this work by adding recent approaches to  explicitly enforce feature compactness by constraining features to lie on a hypersphere  \cite{wang2017normface}
and to have a margin between other  features \cite{deng2019arcface}.    

Fixed classifiers have been recently used also for not discriminative purposes. The work \cite{sablayrolles2018spreading} trains a neural network in which the last layer has fixed parameters with pre-defined points of a hyper-sphere (i.e. a spherical lattice). The work aims at learning a function to build an index that maps real-valued vectors to a uniform distribution over a $d$-dimensional sphere to preserve the neighborhood structure in the input space while best covering the output space. The learned function is used to make high-dimensional indexing more accurate.

\textbf{Softmax Angular Optimization}. Some papers train DCNNs by direct angle optimization \cite{Liu2017CVPR,Liu_2018_CVPR,wang2017normface}. From a semantic point of view, the angle encodes the required discriminative information for class recognition. The wider the angles the better the classes are separated from each other and, accordingly, their representation is more discriminative. The common idea of these works is that of constraining the features and/or the classifier weights to be unit normalized. 
The works \cite{liu_2017_coco_v1}, \cite{hasnat2017mises} and \cite{wang2017normface} normalize both features and weights, while the work \cite{ranjan2017l2} normalizes the features only and \cite{Liu2017CVPR} normalizes the weights only. Specifically, \cite{ranjan2017l2} also proposes adding a scale parameter after feature normalization based on the property that increasing the norm of samples can decrease the Softmax loss \cite{yuan2017feature}. 

From a statistical point of view, normalizing weights and features is equivalent to considering features distributed on the unit hypersphere according to the von Mises-Fisher distribution \cite{hasnat2017mises} with a common concentration parameter (i.e. features of each class have the same compactness). Under this model each class weight represents the mean of its corresponding features and the scalar factor (i.e. the concentration parameter) is inversely proportional to their standard deviations. Several methods implicitly follow this statistical interpretation in which the weights act as a summarizer or as parameterized prototype of the features of each  class \cite{wang2017normface, Liu2017CVPR, wang2018additive, Imprinted_Qi_2018_CVPR, Wu_2018_ECCV}. Eventually, as conjectured in \cite{wang2017normface} if all classes are well-separated, they will roughly correspond to the means of features in each class. 

In \cite{pernici2019fix} the fixed classifiers based on regular polytopes produce features exactly centered around their fixed weights as the training process advances. The work globally imposes the largest angular distances between the class features before starting the learning process without an explicit optimization of the classifier or the requirement of an auxiliary loss as in \cite{HypersphericalEnergy2018} and \cite{kaizhao2019regular}. 
The works \cite{HypersphericalEnergy2018,kaizhao2019regular} add a regularization loss to specifically force the classifier weights during training to be far from each other in a global manner. 
These works including \cite{pernici2019fix} draw inspiration from a well-known problem in physics -- the Thomson problem \cite{thomson1904xxiv} -- where given $K$ charges confined to the surface of a sphere, one seeks to find an arrangement of the charges which minimizes the total electrostatic energy. Electrostatic force repels charges each other inversely proportional to their mutual distance. In \cite{HypersphericalEnergy2018} and \cite{kaizhao2019regular} global equiangular features are obtained by adding to the standard categorical Cross-Entropy loss a further loss inspired by the Thomson problem while \cite{pernici2019fix} builds directly an arrangement for global separability and compactness by considering that minimal energies are often concomitant with special geometric configurations of charges that recall the geometry of regular polytopes in high dimensional spaces \cite{batle2016generalized}.

\section{Regular Polytope Networks with Additive Angular Margin Loss}
In Neural Networks the representation for an input sample is the feature vector $\mathbf{f}$ generated by the penultimate layer, while the last layer (i.e. the classifier) outputs score values according to the inner product as: 
\begin{equation}
{
z_i = \mathbf{w}_i^\top \cdot \mathbf{f}
}
\label{eq_logit}
\end{equation}
for each class $i$, where $\mathbf{w}_i$ is the weight vector of the classifier for the class $i$. In the final loss, the scores are further normalized into probabilities via the Softmax function.

Since the values of $z_i$ can be also expressed as ${z_i = \mathbf{w}^\top_i \cdot \mathbf{f} = ||\mathbf{w}_i|| \: ||\mathbf{f}|| \cos(\theta)}$, where $\theta$ is the angle between $\mathbf{w}_i$ and $\mathbf{f}$, the score for the correct label with respect to the other labels is obtained by optimizing $||\mathbf{w}_i||$, $||\mathbf{f}||$ and $\theta$. 
According to this, feature vector directions and weight vector directions align simultaneously with each other at training time so that their average angle is made as small as possible.
\begin{figure*}[t]
\hspace{-0.3cm}
\includegraphics[width=2.11\columnwidth]{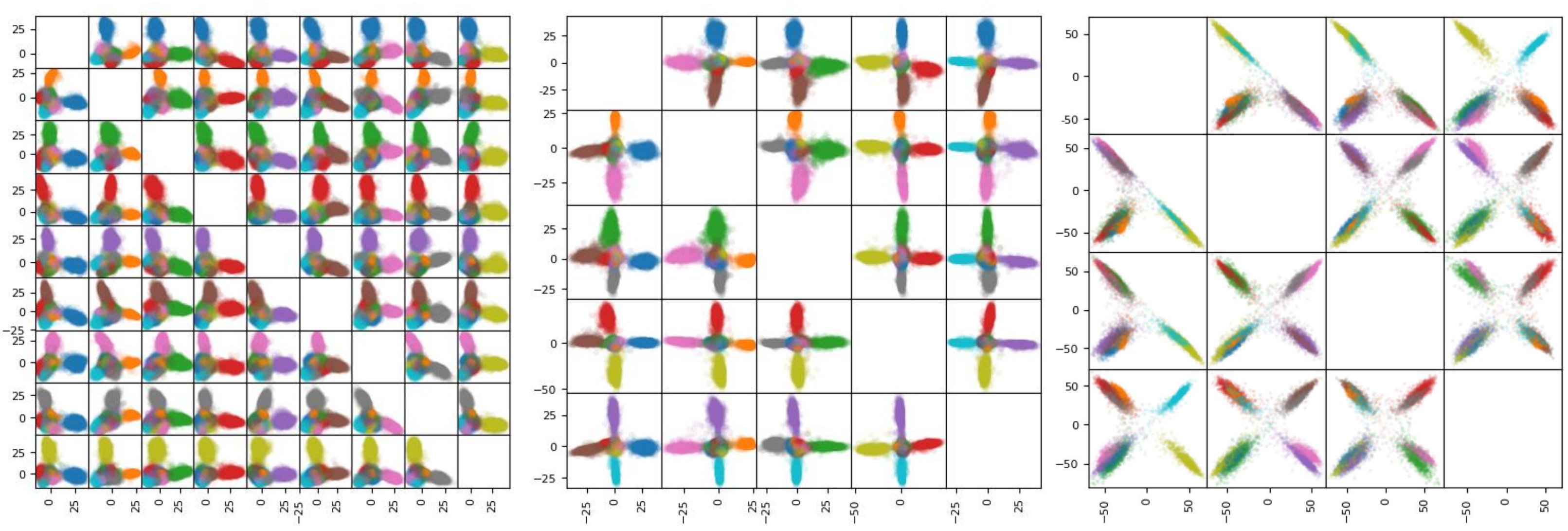}
\caption{The distribution of features learned from the MNIST dataset using the RePoNet classifiers. Features are shown (from left to right) with a scatter plot matrix for the $d$-Simplex, $d$-Orthoplex and $d$-Cube classifier respectively. 
It can be noticed that features are distributed following the symmetric vertex configuration of polytopes. Although features are maximally separated, their compactness is limited. }
\label{fig:limitedcompactness}
\end{figure*}
In \cite{pernici2019fix} it is shown that if classifier weights are excluded from learning, they can be regarded as fixed angular references to which features align. In particular, if the fixed weights are derived from the three regular polytopes available in $\mathbb{R}^d$ with $d \geq 5$, then their symmetry creates angular references to which class features centrally align.
More formally, let $\mathbf{X} = \{(x_i, y_i)\}_{i=1}^{N}$ be the training set containing $N$ samples, where $x_i$ is the image input to the CNN and $y_{i} \in \{1,2,\cdots,K\}$ is the label of the class that supervises the output of the DCNN. Then, the Cross Entropy loss can be written as:
\begin{align}
\mathcal{L}=-\frac{1}{N}\sum_{i=1}^{N} \log\Bigg( \frac {e^{{\mathbf{w}_{y_i}^{\top}\mathbf{f}_{i}+\mathbf{b}_{y_i}}}} {\sum_{j=1}^{K}e^{{\mathbf{w}_{j}^{\top}\mathbf{f}_{i} + \mathbf{b}_{j}}}} \Bigg), 
\label{softmax_loss}
\end{align}
where $\mathbf{W}=\{ \mathbf{w}_j \}_{j=1}^{K}$ are the fixed classifier weight vectors for the $K$ classes. 
Only three polytopes exist in every dimensionality and are: the $d$-Simplex, the $d$-Orthoplex and the $d$-Cube from which three classifiers can be defined as  follow:
\begin{align}
\mathbf{W}_s=\Big \{e_1,e_2,\dots,e_{d-1}, \alpha \sum_{i=1}^{d-1} e_i \Big \},
\label{dsimplex}
\end{align}
\begin{align}
\mathbf{W}_o = \{ \pm e_1,\pm e_2,\dots,\pm e_d \},
\label{dorthoplex}
\end{align}
\begin{align} 
\mathbf{W}_c = \left  \{ \mathbf{w} \in \mathbb {R}^{d}:  \left [-\frac{1}{\sqrt{d}},\frac{1}{\sqrt{d}} \right ] ^d \right \},
\label{dcube}
\end{align} 
where $\alpha=\frac{1-\sqrt{d+1}}{d}$ in Eq.\ref{dsimplex} and $e_i$ with $i \in \{1,2, \dots, d-1\}$ in Eqs.\ref{dsimplex} and \ref{dorthoplex} denotes the standard basis in $\mathbb{R}^{d-1}$. The final weights in Eq.\ref{dsimplex} are further shifted about the centroid, the other two are already centered around the origin. 
Such sets of weights represent the vertices of the generalization of the tetrahedron, octahedron and cube respectively, to arbitrary dimension $d$. 
The weights are further unit normalized ($\hat{\mathbf{w}}_j = \frac{\mathbf{w}_j}{||\mathbf{w}_j||}$) and the biases are set to zero ($\mathbf{b}_j=0$). According to this, Eq.~\ref{softmax_loss} simplifies to: 
\begin{align}
\mathcal{L} =
-\frac{1}{N}\sum_{i=1}^{N}\log\Bigg( 
\frac {e^ {  \hat{\mathbf{w}}_{y_i}^{\top} {\mathbf{f}}_{i} }} {\sum_{j=1}^{K} e^{{{\hat{\mathbf{w}}_j^{\top}\hat{\mathbf{f}}_{i}}}}} \Bigg).
\label{softmax_loss_angles}
\end{align}
Although,  Eq.~\ref{softmax_loss_angles} directly optimizes for small angles, only partial intra-class compactness can be enforced. 
Fig.\ref{fig:limitedcompactness} shows (from left to right) the distribution of features learned from the MNIST dataset with the three different classifiers. The features are displayed as a collection of points, each having the activation of one feature coordinate determining the position on the horizontal axis and the value of the other feature coordinate activation determining the position on the vertical axis. All the pairwise scatter plots of the feature activation coordinates are shown and feature classes are color coded. The size of the scatter plot matrices follows the size of the feature dimensionality $d$ of each fixed classifier which can be determined according to the number of classes $K$ as: 
\begin{align}
d = K-1, \qquad 
d = \lceil \log_2(K) \rceil, \qquad
d = \Big \lceil \frac{K}{2}\Big \rceil,
\end{align}
respectively. The scatter plot matrices therefore result in the following dimensions: $9\times9$, $5\times5$ and $4\times4$ respectively.
As evidenced from the figure, the features follow the symmetric and maximally separated vertex configurations of their corresponding polytopes. 
This is due to the fact that each single pairwise scatter plot is basically a parallel projection onto the planes defined by pairs of multidimensional axes. According to this, features assume a $\upY$, $+$, and $\times$ shaped configuration for the $d$-Simplex, $d$-Orthoplex and $d$-Cube respectively.
Although maximal separation is achieved, the intra-class average distance is large and therefore not well suited for recognition purposes.

The plotted features are obtained training the so called LeNet++ architecture \cite{wen2016discriminative}. The network is a modification of the LeNet architecture \cite{lecun1998gradient} to a deeper and wider network including parametric rectifier linear units (pReLU) \cite{he2015delving}.
The network is learned using the Adam optimizer \cite{kingma2014adam} with a learning rate of $0.0005$, the convolutional parameters are initialized following \cite{KaimingHe15} and the mini-batch size is $512$.

To improve compactness keeping the global maximal feature separation we follow \cite{wang2018cosface, wang2017normface}  normalizing the features and multiplying them by a scalar $\kappa$: $\hat{\mathbf{f}}_i = \frac{\mathbf{f}_i}{||\mathbf{f}_i||} \kappa$. The loss in Eq.\ref{softmax_loss} can be therefore rewritten as: 
\begin{align}
\mathcal{L} & =
-\frac{1}{N}\sum_{i=1}^{N}\log\Bigg( 
\frac {e^ { \kappa \hat{\mathbf{w}}_{y_i}^{\top} \hat{\mathbf{f}}_{i} }} {\sum_{j=1}^{K}  e^{{{\kappa \hat{\mathbf{w}}_j^{\top}\hat{\mathbf{f}}_{i}}}}} \Bigg)
\nonumber
\\
& =-\frac{1}{N}\sum_{i=1}^{N}\log\Bigg( 
\frac {e^ { \kappa \cos(\theta_{y_{i}}) }} {\sum_{j=1}^{K} e^{
\kappa \cos(\theta_j)
}} \Bigg)
\label{softmax_loss_von}
\end{align}
The equation above minimizes the angle $\theta_{y_{i}}$ between the fixed weight corresponding to the label $y_{i}$ and its associated feature. The equation can be interpreted as if features are realizations from a set of $K$ von Mises-Fisher distributions having a common concentration parameter $\kappa$. Under this parameterization $\hat{\mathbf{w}}$ is the mean direction on the hypersphere and $\kappa$ is the concentration parameter. The greater the value of $\kappa$ the higher the concentration of the distribution around the mean direction $\hat{\mathbf{w}}$ and the more compact the features. 
This value has already been discussed sufficiently in several previous works \cite{wang2017normface,ranjan2017l2}. In this paper, we directly
fixed it to $30$ and will not discuss its effect anymore.

To obtain maximal compactness the additive angular margin loss described in \cite{deng2019arcface} is exploited. According to this, Eq.\ref{softmax_loss_von} becomes: 
\begin{align}
\mathcal{L}=-\frac{1}{N} \sum_{i=1}^{N} \log \Bigg( 
\frac{e^{\kappa \cos(\theta_{y_{i}}+m)}}
{e^{\kappa \cos(\theta_{y_{i}}+m)}+
\sum\limits_{ \tiny \substack{j=1 \\ j \neq y_{i}}}^{n} e^{\kappa \cos(\theta_{j})}} \Bigg),
\label{margin_loss}
\end{align}
where the scalar value $m$ is
an angle in the normalized hypersphere introducing a margin between class decision boundaries.
The loss of Eq.~\ref{margin_loss} together with the fixed classifier weights of Eqs.~\ref{dsimplex}, \ref{dorthoplex}, \ref{dcube} allows learning discriminative features without using any auxiliary loss other than the Softmax.

\begin{figure}[t]
\centering
\includegraphics[width=0.99\columnwidth]{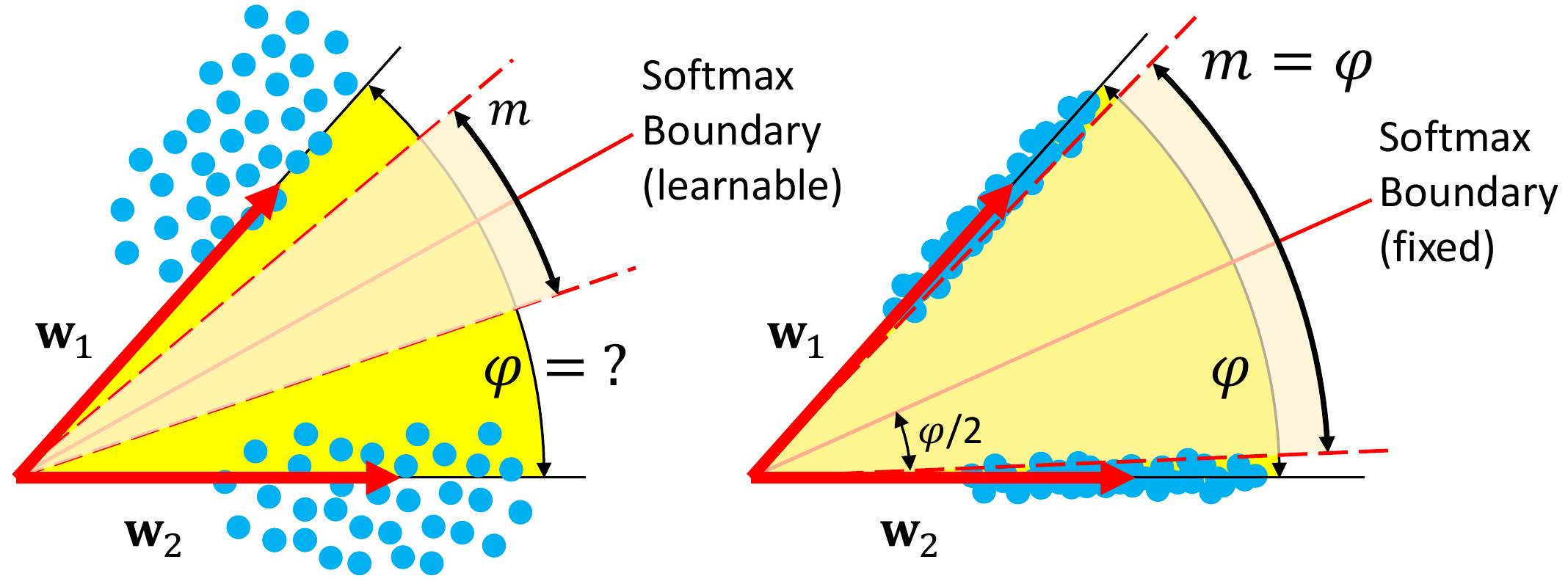}
\caption{ Maximally compact feature learning with \mbox{RePoNet} fixed classifiers and the angular margin loss.
\emph{Left}: In a standard learnable classifier the decision boundaries (dashed lines) defined by the angular margin $m$ do not push features to their respective weights uniformly (red arrows).
\emph{Right}: In RePoNet classifiers the margin can be analytically determined ($m=\varphi$) so that the decision boundaries maximally push the features closer to their respective fixed weight. }
\label{fig:max_compactness}
\end{figure}

The advantage of our formulation is that $m$ is no longer an hyperparameter that have to be searched. Indeed, the loss above when used with RePoNet classifiers is completely interpretable and the margin $m$ can be set according to the angle $\varphi$ subtended between a class weight and its connected class weights as illustrated in Fig.\ref{fig_IntroFig}. 
For each of the three RePoNet fixed classifiers the angle $\varphi$ can be analytically determined as \cite{pernici2019fix}:
\begin{align}
\varphi_s & = \arccos{\bigg(-\frac{1}{d}\bigg)}, 
\label{eq_dsimplex_angle}
\\
\varphi_o & = \frac{\pi}{2},
\label{eq_dortho_angle}
\\
\varphi_c & = \arccos\bigg(\frac{d-2}{d}\bigg), 
\label{eq_dcube_angle}
\end{align}
respectively, where $d$ is the feature space dimension size. 
Fig.~\ref{fig:max_compactness} shows the effect of setting:
$$
m=\varphi.
$$
In the figure we draw
a schematic 2D diagram to show the effect of the margin $m$ on pushing the class decision boundary to achieve feature compactness.
In the standard case of a learnable classifier, as shown in Fig.~\ref{fig:max_compactness}~\emph{(left)}, the value $\varphi$ is not known in advance, it varies from class to class and features are not guaranteed to distribute angularly centered around their corresponding weights. Therefore, $m$ cannot be set in an interpretable way. Contrarily, in the case proposed in this paper and shown in Fig.~\ref{fig:max_compactness}~\emph{(right)}, the value $\varphi$ is constant and known in advance, therefore by setting $m=\varphi$, the class decision boundaries are maximally pushed to compact features around their fixed weights. 
This because the Softmax boundary (from which the margin is added) is exactly in between the two weights $\mathbf{w}_1$ and $\mathbf{w}_2$. 
According to this, the features generated by the proposed method are not only \emph{maximally separated} but also \emph{maximally compact} (i.e. maximally discriminative). 

\section{Exploratory Results} 
Experiments are conducted with the well-known MNIST and EMNIST \cite{EMNIST17} datasets. 
MNIST contains $50,000$ training images and $10,000$ test images. The images are in grayscale and the size of each image is $28 \times 28$ pixels. There are 10 possible classes of digits. The EMNIST dataset (balanced split) holds $112,800$ training images, $18,800$ test images and has $47$ classes including lower/upper case letters and digits.

\begin{figure}[t]
\centering
\includegraphics[width=0.99\columnwidth]{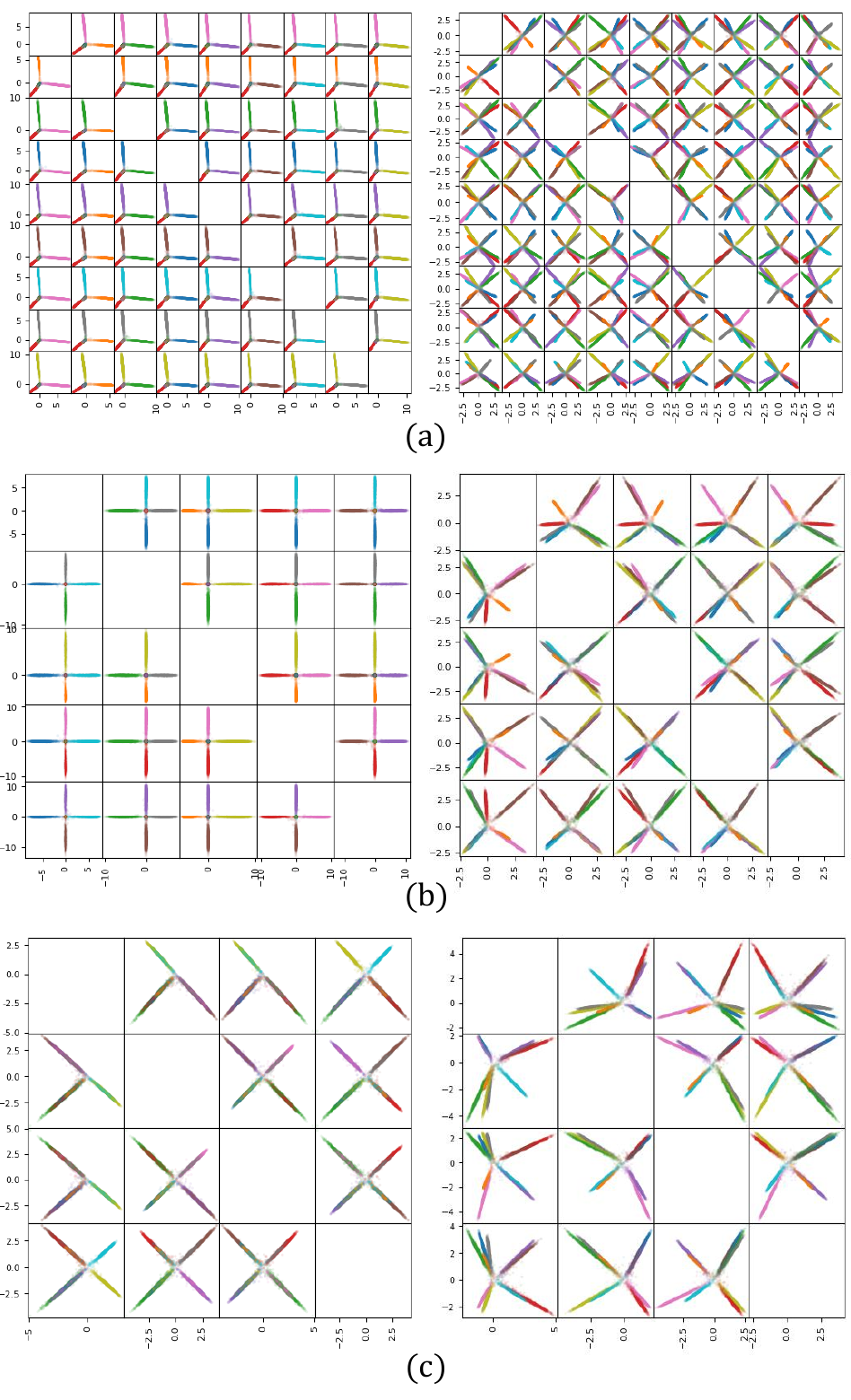}
\caption{The distribution of MNIST learned features using the proposed method (\emph{Left}) and learned using a  standard trainable classifier (\emph{Right}). The scatter plot highlights the maximal separability and compactness of the extracted features for the (a) $d$-Simplex, (b) $d$-Orthoplex and (c) $d$-Cube classifiers. Class features are color coded. As the feature space dimension decreases standard baselines have difficulty in obtaining inter-class separation.}
\label{fig:scatter_mnist}
\end{figure}

Fig.~\ref{fig:scatter_mnist} shows a visual comparison between the features generated by the RePoNet fixed classifiers (\emph{left column}) and by a standard CNN baseline with learnable classifiers (\emph{right column}). Both approaches are trained according to the loss of Eq.~\ref{margin_loss} and have exactly the same architecture, training settings and embedding feature dimension used in  Fig.~\ref{fig:limitedcompactness}. Results are presented with a scatter plot matrix. Although the two methods achieve substantially the same classification accuracy (i.e. $99.45\%$ and $99.47\%$ respectively), it can be noticed that the learned features are different. Specifically, Margin-RePoNet follows the exact configuration geometry of their related polytopes. Features follow very precisely their relative $\upY$, $+$, and $\times$ shapes therefore achieving maximal separability. The standard baselines with learnable classifiers (Fig.~\ref{fig:scatter_mnist} \emph{left column}) achieve good but non maximal separation between features. However, as the embedding dimension decreases, as in Fig.~\ref{fig:scatter_mnist}(c), the separation worsens. 

\begin{figure}[t]
\centering
\includegraphics[width=0.99\columnwidth]{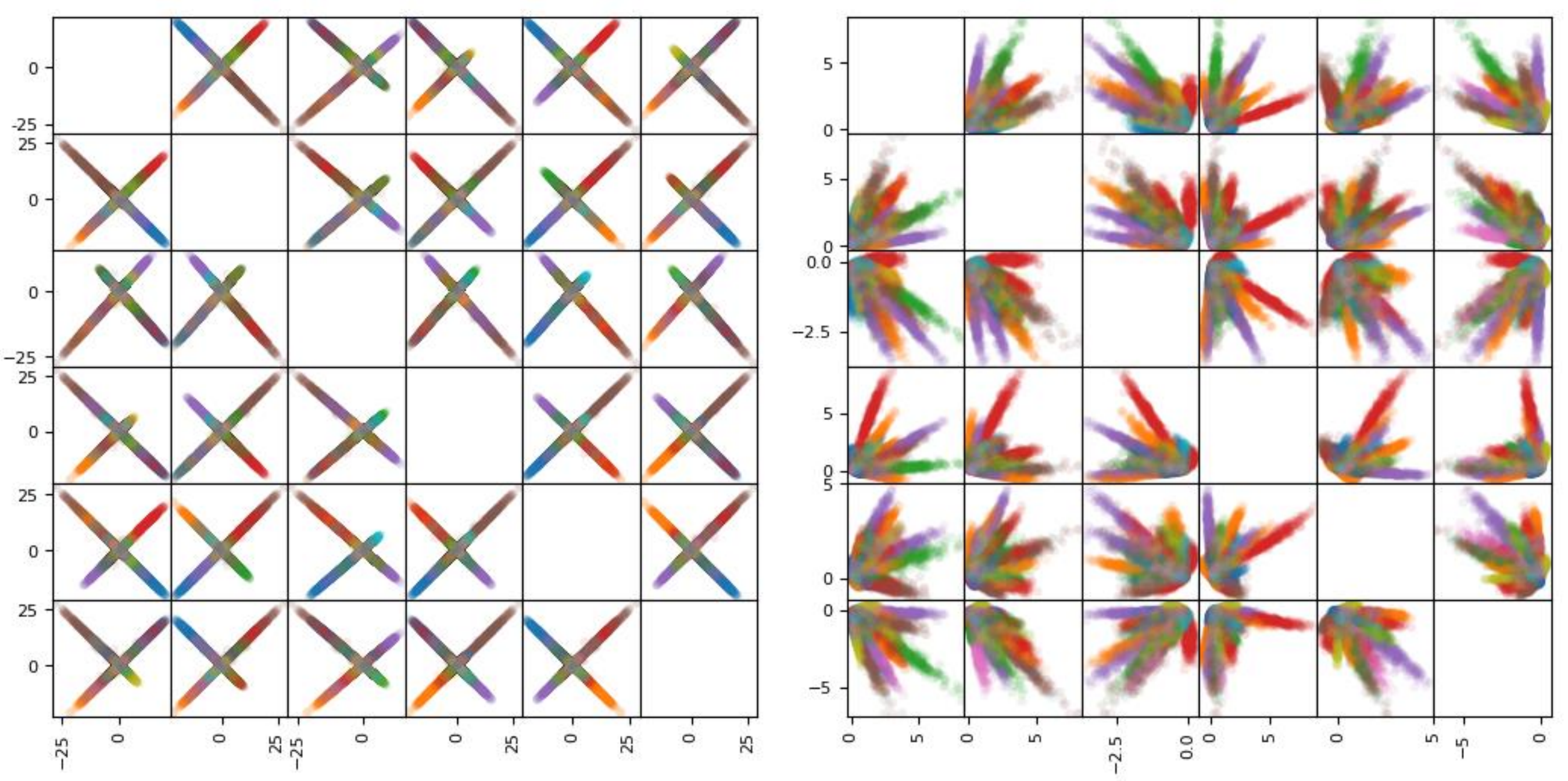}
\caption{The distribution of EMNIST (balanced split) learned features. \emph{Left}: Features learned using the $d$-Cube fixed classifier with the additive angular margin loss. \emph{Right}: Features learned using a standard trainable classifier with the additive angular margin loss. In both cases the feature dimension is $6$ and the classification accuracy is comparable. Maximal separability and compactness are evident in our approach. }
\label{fig:scatter_EMNIST}
\end{figure}

This effect is particularly evident in more difficult datasets. Fig.\ref{fig:scatter_EMNIST} shows the same visual comparison using the \mbox{EMNIST} dataset where some of the $47$ classes are difficult to be correctly classified due to their inherent  ambiguity.
Fig.~\ref{fig:scatter_EMNIST} shows the scatter plot matrix of the $d$-Cube classifier (\emph{left}) compared with its learnable classifier baseline (\emph{right}) in dimension $d=6$. Although also in this case they both achieved the same classification accuracy (i.e. $88.31\%$ and $88.39\%$), the features learned by the baseline are neither well separated nor compact.

\begin{figure}[t]
\centering
\includegraphics[width=0.99\columnwidth]{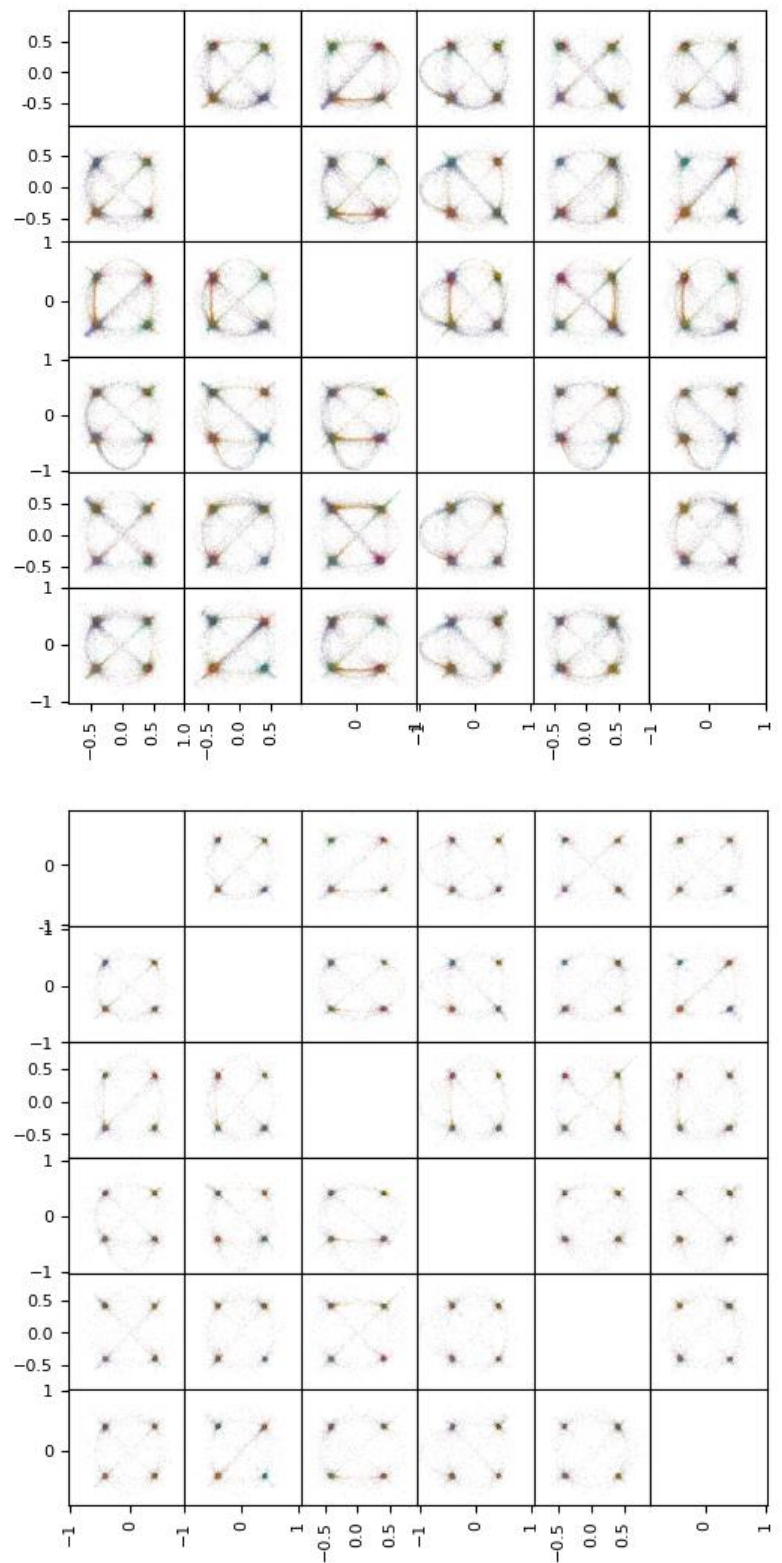}
\caption{The distribution of the EMNIST \emph{normalized} learned features shown in \ref{fig:scatter_EMNIST}~\emph{(Left)}. (\emph{Top}) training-set. (\emph{Bottom}) test-set (best viewed in electronic version). }
\label{fig:scatter_EMNIST_norm}
\end{figure}

Finally, in Fig.~\ref{fig:scatter_EMNIST_norm} we show the $L_2$ normalized features (typically used in recognition) of both the training (\emph{top}) and test set (\emph{bottom}) for the same experiment shown in Fig.~\ref{fig:scatter_EMNIST}. 
Class features in this case correctly follow the vertices of the six-dimensional hypercube since all the parallel projections defined by each pairwise scatter plot result in the same unit square centered at the origin.

\section{Conclusion}
We have shown how to extract features from Convolutional Neural Networks with the desirable properties of maximal separation and maximal compactness in a global sense. We used a set of fixed classifiers based on regular polytopes and the additive angular margin loss. The proposed method is very simple to implement and preliminary exploratory results are promising.

Further implications may be expected in large face recognition datasets with thousands of classes (as in \cite{Cao18}) to obtain maximally discriminative features with a significant reduction in: the number of model parameters, the feature size and the hyperparameters to be searched.
\bibliographystyle{unsrt}
\bibliography{egbib}

\end{document}